# Parallel Computing Architectures for Robotic Applications: A Comprehensive Review

Md Rafid Islam
Electrical and Computer Engineering
North South University
Dhaka, Bangladesh
md.islam.241@northsouth.edu

***Abstract—****With the growing complexity and capability of contemporary robotic systems, the necessity of sophisticated computing solutions to efficiently handle tasks such as real-time processing, sensor integration, decision-making, and control algorithms is also increasing. Conventional serial computing frequently fails to meet these requirements, underscoring the necessity for high-performance computing alternatives. Parallel computing, the utilization of several processing elements simultaneously to solve computational problems, offers a possible answer. Various parallel computing designs, such as multi-core CPUs, GPUs, FPGAs, and distributed systems, provide substantial enhancements in processing capacity and efficiency. By utilizing these architectures, robotic systems can attain improved performance in functionalities such as real-time image processing, sensor fusion, and path planning. The transformative potential of parallel computing architectures in advancing robotic technology has been underscored, real-life case studies of these architectures in the robotics field have been discussed, and comparisons are presented. Challenges pertaining to these architectures have been explored, and possible solutions have been mentioned for further research and enhancement of the robotic applications.*

## 1. Introduction

Robotics is a field of technology that encompasses aspects of designing, constructing, operating, and utilizing robots. It combines elements of mechanical engineering, electrical engineering, computer science, and related fields to develop machines capable of assisting, enhancing, or substituting human tasks [1]. Robotics has seen fast advancement, becoming an essential part of diverse fields such as manufacturing, healthcare, autonomous cars, and space exploration. Robots play an important role in automating repetitive tasks, leading to increased efficiency, and they can aid in quality control by performing high-precision inspections. Robotics has played an important role in the medical industry and healthcare system. For example, the da Vinci Surgical System facilitates minimally invasive procedures with high precision, resulting in shorter recovery times and improving patient outcomes [2]. Robotic technologies like the Mars rovers Curiosity and Perseverance have been used in planetary exploration, experiments, and data collection in hostile environments [3]. Robotics has also been instrumental in developing autonomous vehicles, providing mobility for non-drivers, and improving logistics [4].

A range of computational challenges occur when utilizing robots due to the diversity and complexity of the tasks they carry out. Latency in dynamic environments can even lead to accidents. The reliance of robots on various sensors for environmental perception and data synchronization from a variety of diverse sources to form an understanding of their surroundings is a complex computational task.

Determining optimal paths in real time and making autonomous decisions require significant computational power [5]. Implementation of control algorithms for precision, stability, and adaptability of robotic systems is also computationally intensive [6].

Traditional serial computing in robotics has a number of limitations. Sequential execution of tasks leads to bottlenecks when handling large datasets. It is difficult with serial execution to take decisions based on sensor data processing in milliseconds, and it is unable to take advantage of parallelism when tasks like image processing or sensor fusion are inherently parallel. Modern robots process vast amounts of data, which serial computation cannot handle due to its complexity. With robots' growing sophistication and increasing computational needs, serial computers fail to scale appropriately [7].

To meet these computational challenges, High-Performance Computing (HPC) plays a critical role by enabling parallel processing, the use of deep learning for perception, and reinforcement learning for decision-making [8]. HPC's usage of multiple processors to execute tasks in parallel increases the processing power, which is crucial for real-time image processing. HPC systems are scalable and adaptable based on the needs of specific robotic applications and are capable of meeting the real-time data handling requirements of modern robots by processing large amounts of data with minimal latency [9]. Furthermore, HPC allows for the use of advanced machine learning and AI algorithms, and it is capable of handling the intensive computations required by these algorithms.

Parallel computing involves the simultaneous execution of numerous processes, the division of large problems into smaller ones for concurrent solutions, and the utilization of multiple processors [10]. Parallel computing architecture is the design and organization of a system that facilitates multiple processors to work in unison to solve a problem simultaneously and leads to substantial improvements in computational efficiency [11]. Through the utilization of parallel computing, image frames can be divided into smaller pieces, and multiple processors can process them simultaneously. Parallel algorithms can achieve complex data processing and real-time synchronization, which are necessary for a coherent understanding of the environment. Robots can navigate more quickly in a dynamic environment by parallelizing path planning algorithms such as Rapidly Exploring Random Trees (RRT) or A*, which explore multiple branches concurrently [12]. Parallel computing can provide a solution to implementing advanced control algorithms, like Model Predictive Control (MPC), to deal with real-time constraints [13].

Modern multi-core CPUs are able to execute several threads in parallel, and GPUs are equipped with a large number of cores that are specifically designed to handle parallel processing workloads. Combining CPUs and GPUs in a single system provides a balanced approach, as CPUs handle serial processing efficiently while GPUs speed up parallel tasks. The distributed approach of high-performance computing clusters scales efficiently, and services like AWS, Azure, and Google Cloud enable powerful computing resource usage without significant upfront investments through scalable HPC solutions [14]. Neuromorphic computing architecture shows great potential for AI and machine learning in robotics, as it emulates the brain's functionality in decision-making and pattern recognition [15].

The intersection of parallel computing architecture and robotics is a rapidly evolving field, and the integration of these two domains is inherently interdisciplinary, which introduces complexity that requires the synthesization of knowledge from diverse fields. While there are reviews available, they do provide a holistic view of the subject matter, and given the pace of the advancements, reviews a couple of years old may be missing in recent advancements and thus prove to be outdated. Therefore, there is a need for a comprehensive review with the goal of bridging the gaps in existing literature and highlighting directions for future research.

The remainder of this paper is organized as follows: Related works are reviewed in Section 2, parallel computing architecture types and robotic applications are discussed in Sections 3 and 4, case studies are presented in Section 5, comparisons among architectures are made in Section 6, challenges and potential solutions are explored in Section 7, and the paper ends with a discussion and conclusion in Sections 8 and 9.

## 2. Background

### 2.1 Milestones in Parallel Computing

Parallel computing has evolved significantly from its early theoretical concepts in the 1950s, where John von Neumann's designs hinted at parallel processing. In the 1960s, Seymour Cray's CDC 6600 introduced the idea of multiple functional units for parallel processing. The 1970s saw the development of the ILLIAC IV, one of the first large-scale parallel computers, and the Cray-1, which featured vector processing. The 1980s marked the expansion of parallel computing with the Connection Machine's massively parallel architecture and the introduction of Intel's iPSC/1. The development of the Message Passing Interface (MPI) in 1986 standardized processor communication. By the 1990s, IBM's SP1 and SP2 systems and the Beowulf Project had popularized parallel computing clusters using commodity hardware. The 2000s brought multi-core processors from Intel and AMD, along with NVIDIA's CUDA platform, revolutionizing GPU-based parallel computing. OpenCL in 2011 further enabled programming across heterogeneous platforms. Major milestones continued with China's Tianhe-2 and Japan's Fugaku supercomputers, showcasing advances in parallel computing power. The ongoing development of quantum computing promises to further revolutionize parallel processing by leveraging qubits for exponential speedups in solving complex problems [16].

### 2.2 Types of Parallel computing

The three primary types of parallel computing are data parallelism, task parallelism, and pipeline parallelism. Data parallelism refers to the distribution of subsets of identical data among several processors, with each processor performing the same action on its assigned portion simultaneously. Task parallelism, also referred to as functional parallelism, partitions the computation into distinct tasks for concurrent execution. Each processor performs a different task, often operating on the same or different data. Pipeline parallelism, also known as pipelining, involves dividing a task into a sequence of consecutive stages, where each stage is handled by a distinct processor. A sequential flow of processing steps is established by passing the output of one stage as the input to the next stage [17, 18].

### 2.3 Principles of Parallel Computing

Parallel computing's core principles include decomposing a computer problem into smaller components, concurrency, task coordination and synchronization, parallel system scalability, and load balancing [19]. Decomposition involves dividing a problem into discrete components and dividing data into smaller segments, allowing for independent task execution and parallel data processing. Multi-threading achieves concurrent execution by executing tasks at overlapping times. Locks, semaphores, barriers, and atomic operations coordinate access to shared resources for synchronization, and message-passing shared memory access or parallel communication libraries like MPI enable communication between parallel tasks. Strong scalability is the ability to achieve faster execution with more processors while keeping the problem size constant. Weak scalability aims to maintain the execution time as the problem size and the number of processors grow in proportion. Load balancing ensures the optimal use of all processors, preventing both idleness and overload. Redundancy of tasks and periodic computational state-saving ensure fault tolerance. Finally, identifying parallelism, minimizing dependencies, and optimizing communication are of utmost importance for parallel algorithm design [20].

### 2.4 Robotics and Computational Demands

The computational demands of modern robots are multifaceted and encompass real-time processing, sensor integration, path planning, machine learning, and control systems [21]. Operating in real-time environments with obstacle avoidance and motion control requires low latency. High-frequency data generated by the likes of LIDAR and IMUs must be processed rapidly for real-time decision-making. Generally, techniques like Kalman filtering, particle filtering, and deep learning-based approaches are used to integrate data from various sensors to form a coherent picture, which requires significant computational resources. Training and deploying deep learning models for robotic use entails effectively running complex neural networks. Path planning and optimal path calculation use algorithms like A*, D*, and RRT (Rapidly Exploring Random Tree), and their variants are computationally intensive, especially in dynamic and uncertain environments, to account for changes in the environment that require re-computation and path adjustment [22, 23]. Implementing robotic control systems requires continuous computational effort that uses feedback loops for stability and achieves desired behaviors, such as PID (Proportional-Integral-Derivative) controllers, Model Predictive Control (MPC), or more advanced adaptive learning-based controllers [24, 25]. Kinematics and dynamics calculations, high fidelity simulations, and virtual prototyping for testing and validation require extensive computational power. In the case of a multi-robot system, ensuring effective communication and managing latency and bandwidth are crucial, and they also add to the already existing computational complexity.

### 2.5 Current Limitations and Challenges

Notwithstanding the progress in robotics, the computational requirements of contemporary robotic systems are well beyond the capabilities of conventional serial computing techniques. Large-scale data processing and the effective implementation of intricate algorithms are the causes of performance bottlenecks. This restriction reduces robot performance generally and their real-time capabilities. Another significant issue is energy efficiency, specifically for battery-powered and mobile robotic systems. The power consumption of traditional computational systems sometimes restricts the effectiveness and duration of these robots [26]. The enhanced autonomy and accessibility of robotic systems across various applications rely on energy-efficient computation. Additionally, scalability is of great significance. The expansion of advanced robotic applications may be too much for traditional computing systems to manage; therefore, scalable alternatives must be developed [27].

## 3. Types of Parallel Computing Architectures

### 3.1 Multi-Core Processors

Multi-core processors have an architecture that combines numerous processing units into a single integrated circuit. Each core functions autonomously, with the ability to carry out instructions, and is usually equipped with its own execution unit, registers, and an L1 cache for rapid retrieval of frequently accessed data. Multi-core CPUs employ many levels of cache memory to optimize performance. The L1 cache is exclusive to each core, however, the L2 and occasionally L3 caches can be either shared among cores or exclusive, depending on the design, in order to optimize both performance and efficiency. Inter-core communication is enabled through an interconnect or bus system, providing efficient data transport and coordination [28].

The memory controller is a crucial element in multi-core design as it oversees the transfer of data between the cores and the main memory (RAM), ensuring efficient memory access and minimizing performance limitations. The design additionally has input/output (I/O) interfaces responsible for managing connectivity with external devices, including storage drives, network interfaces, and graphics cards. Multi-core processors generally adhere to a Symmetric Multiprocessing Model (SMP), in which each core has equitable access to memory and I/O resources, facilitating even allocation of workloads.

Cache coherence methods, such as MESI (Modified, Exclusive, Shared, Invalid), are used to preserve data consistency among various caches, guaranteeing that all cores possess a consistent perspective on memory. The processor is equipped with advanced power management features that enable it to adapt power consumption and performance in real-time, depending on the current workload. This allows for the optimization of energy efficiency and heat output. Multi-core processors can achieve improved performance for multi-threaded programs and efficient multitasking by utilizing thread-level parallelism (TLP) to concurrently run numerous threads across different cores. This architectural design offers a flexible and effective way to fulfill the requirements of contemporary computing environments, ranging from personal devices to powerful servers [29].

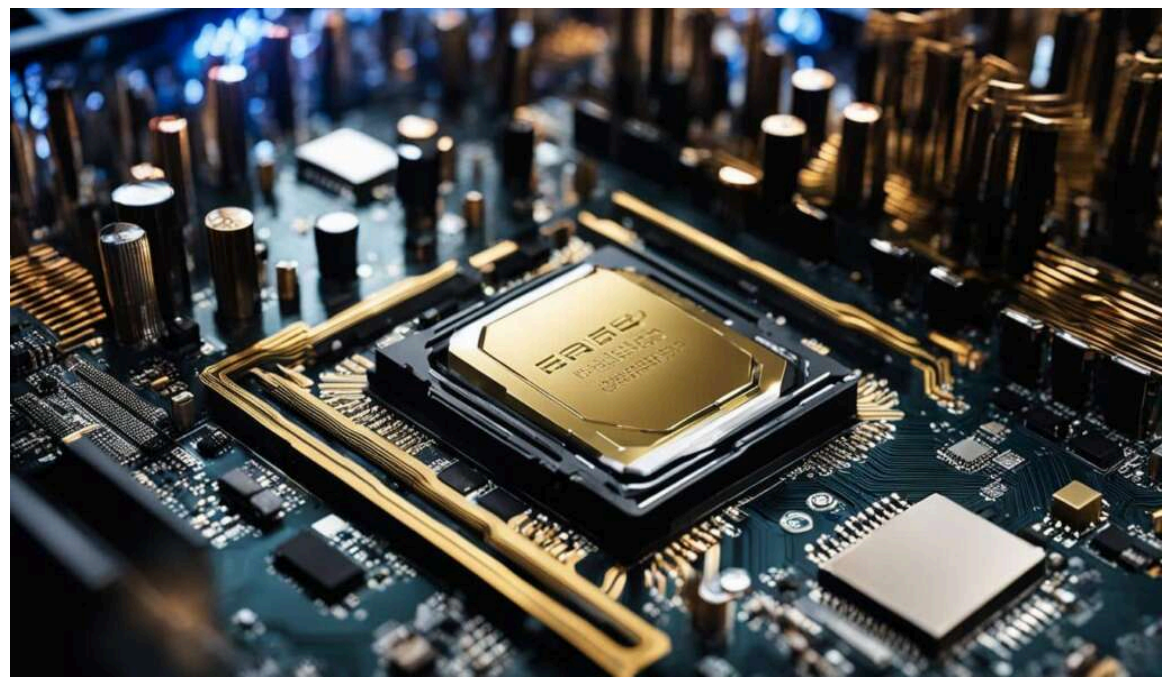

*Figure 1: Multi-Core Processor*

**3.2 Graphics Processing Units (GPUs)**

Graphics Processing Units (GPUs) are dedicated hardware specifically engineered to enhance the speed and efficiency of rendering and parallel processing operations. The GPU architecture is distinguished by a multitude of smaller, highly efficient cores that are specifically designed to do several tasks concurrently. The cores are arranged in clusters, commonly known as Streaming Multiprocessors (SMs) in NVIDIA GPUs or Compute Units (CUs) in AMD GPUs. Each SM or CU is equipped with several arithmetic logic units (ALUs), registers, and a limited amount of local memory, enabling them to execute multiple operations simultaneously [30].

GPUs utilize a hierarchical memory system in order to facilitate efficient data processing with high throughput. This system comprises three basic components: global memory, which serves as the primary memory accessible by all cores; shared memory, which facilitates efficient data sharing inside a single SM or CU; and many layers of cache, including L1 and L2 caches, that enable faster retrieval of frequently accessed data. Optimizing memory utilization is essential for speed due to the increased latency of global memory compared to shared memory and caches.

The design additionally integrates dedicated hardware components for certain functions, such as texture mapping units (TMUs) for managing texture data and render output units (ROPs) for handling final pixel data. Contemporary graphics processing units (GPUs) incorporate supplementary functionalities such as tensor cores, specifically engineered to expedite machine learning tasks, and ray tracing cores, which augment real-time rendering capabilities.

An essential element of GPU architecture is its exceptional memory bandwidth, which is achieved through the use of wide memory interfaces and advanced memory technologies such as GDDR6 or HBM (High Bandwidth Memory). GPUs have the ability to efficiently transfer substantial quantities of data, which is crucial for applications such as graphics rendering and scientific computations [31].

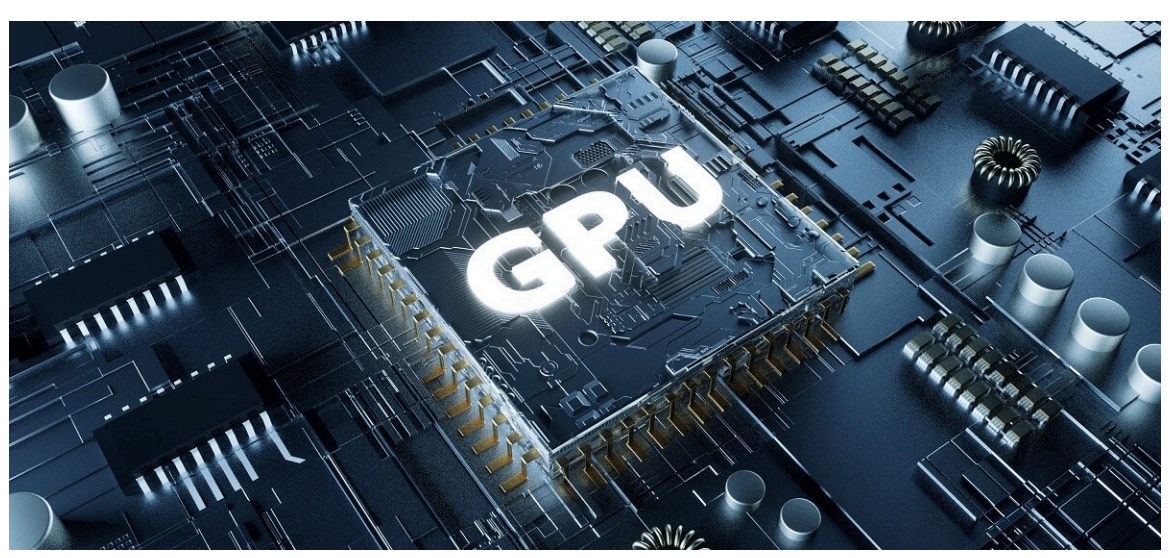

*Figure 2: Graphic Processing Unit*

**3.3 Field-Programmable Gate Arrays (FPGAs)**

Field-Programmable Gate Arrays (FPGAs) are programmable semiconductor devices that can be modified after manufacturing to fulfill various jobs. An FPGA's design consists of an array of programmable logic blocks that can be programmed to execute intricate combinational and sequential logic functions [32]. The logic blocks are interconnected using a programmable interconnect network, enabling them to be joined in almost any desired arrangement. The FPGAs' flexibility allows it to be customized for specialized computing workloads or to simulate different hardware circuits.

The fundamental component of each logic block is a look-up table (LUT), which can be configured to execute any Boolean function. Flip-flops or registers, are included with the LUTs to store the output of the logic blocks and enable the processing of sequential logic and state machines. In addition, Field-Programmable Gate Arrays (FPGAs) incorporate specialized memory blocks known as block RAM (BRAM), which offer rapid, on-chip storage for both data and instructions. The memory blocks can be customized in different dimensions and configurations to meet the requirements of the application.

The architecture has a diverse array of input/output (I/O) blocks that provide various communication protocols and interfaces, enabling the FPGA to establish connections with other hardware components and systems. FPGAs are particularly versatile in interacting with numerous external devices since these I/O blocks can be programmed to handle multiple voltage levels, speeds, and signal standards.

Hardware description languages (HDLs) like VHDL and Verilog are employed to specify the behavior and structure of the FPGA's logic, in order to effectively handle the intricacy of programming an FPGA. In addition, high-level synthesis (HLS) tools enable developers to express algorithms using higher-level programming languages like C/C++, which are subsequently transformed into FPGA configurations through compilation [33].

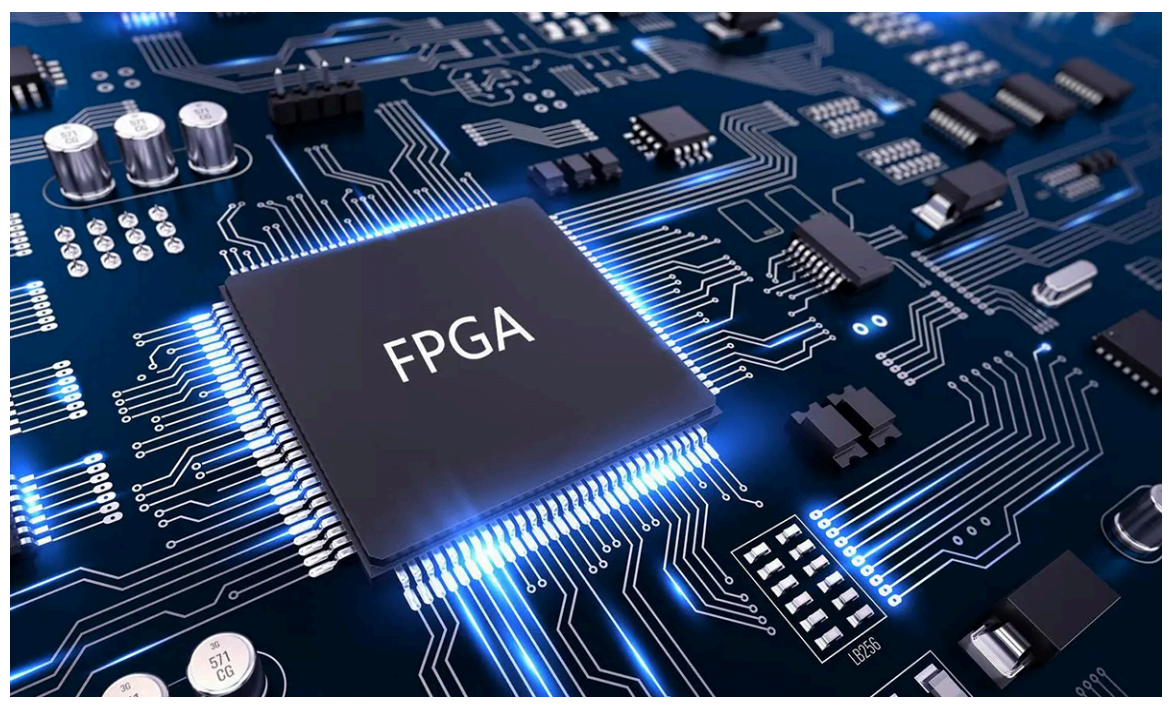

*Figure 3: Field-Programmable Gate Arrays*

### 3.4 Distributed Systems and Cloud Computing

Distributed systems and cloud computing architectures are specifically engineered to offer scalable, dependable, and efficient computing resources by harnessing a network of networked computers. A distributed system is a network of autonomous computing nodes that collaborate to accomplish a shared objective, frequently across different geographic locations. These nodes engage in communication and synchronization by exchanging messages via a network, which may be a local area network (LAN) or a wide area network (WAN). Each node within a distributed system possesses its own distinct local memory and computational skills, which allow for parallel processing and the ability to withstand faults. This means that if one node fails, it does not necessarily result in the complete failure of the entire system.

Distributed systems architecture typically comprises three essential components: computational nodes, communication infrastructure, and a software layer responsible for coordinating and facilitating communication among the nodes. The computational nodes refer to discrete computers or servers, each possessing the ability to execute tasks and retain data. The communication infrastructure comprises networking hardware and protocols that facilitate the exchange of data between nodes. The software layer commonly incorporates middleware, which simplifies the intricacy of the underlying hardware and offers functions such as load balancing, fault tolerance, and data replication [34].

Cloud computing is commonly divided into three service models: Infrastructure as a Service (IaaS), which offers virtualized computing resources via the internet; Platform as a Service (PaaS), which provides a platform for customers to develop, run, and manage applications without needing to handle the underlying infrastructure; and Software as a Service (SaaS), which delivers software applications over the internet through a subscription model.

The fundamental elements of cloud computing consist of the front-end interface, which users engage with through web browsers or applications; the back-end infrastructure, which encompasses servers, storage, and networking equipment; and the management layer, which coordinates resource allocation, load balancing, and security. Advanced cloud designs include containerization with microservices, enabling modular and scalable deployment of applications [35].

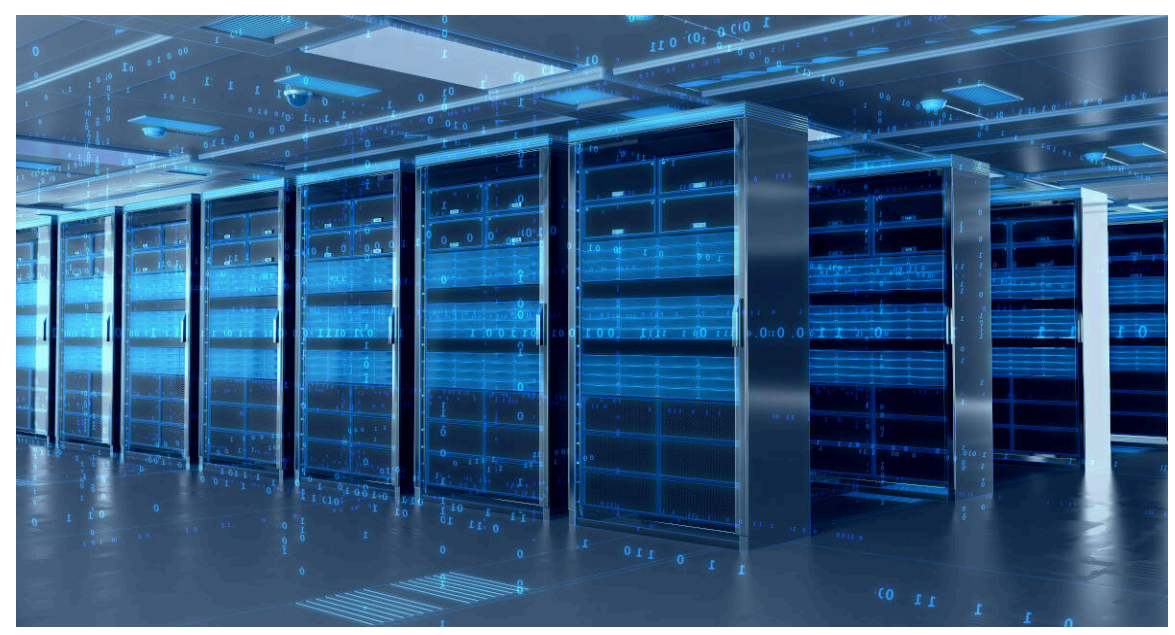

*Figure 4: Distributed Systems*

## 4. Parallel Computing in Robotic Applications

### 4.1 Real-Time Image Processing

Real-time image processing is crucial in various robotic applications, which include autonomous navigation, object recognition, and quality inspection in industrial settings. A multi-core processor performs filtering, edge detection, and feature extraction simultaneously at the various stages of image processing pipelines. The most crucial point for utilizing GPUs in this task is a high throughput with parallelism that will enable quick execution of operations—e.g., convolution operations in neural networks. FPGAs offer parallelism that can be tailored to fit image-processing tasks, thus leading to the implementation of dedicated hardware accelerators that provide better latency and power [36].

### 4.2 Sensor Fusion

Sensor fusion involves combining data from multiple sensors to create a more accurate and reliable representation of the environment, which is critical for robotic tasks such as localization, mapping, and navigation. Multi-core processors can parallelize the processing of different sensor data streams, improving the efficiency of fusion algorithms, and GPUs can accelerate the calculation of sensor fusion algorithms involving matrix operations and statistical methods, while distributed architectures can handle sensor data from multiple robots, enabling cooperative sensing and decision-making in large-scale robotic systems [37].

### 4.3 Path Planning and Navigation

Path planning and navigation are fundamental to autonomous robotic systems, requiring the computation of safe and efficient paths in dynamic environments. Multi-core architectures can parallelize the computation of potential paths and the evaluation of their feasibility, speeding up the planning process, and GPUs can accelerate graph-based algorithms and sampling-based methods used in path planning, such as Rapidly-exploring Random Trees (RRT) and Probabilistic Roadmaps (PRM). Custom hardware can be implemented by FPGAs for real-time path planning and collision detection, providing low-latency solutions that are appropriate for high-speed applications [38].

### 4.4 Control Algorithms

Control algorithms fundamentally govern robotic operations, regulating robots' movement and interaction with their surroundings. Multi-core processors can run complex control algorithms that require real-time processing, such as model predictive control and adaptive control; GPUs can handle the parallel computation of control laws for multiple degrees of freedom in robotic manipulators, enhancing the performance of tasks requiring precise motion control; and FPGAs provide deterministic and low-latency execution of control algorithms, making them ideal for applications requiring high reliability and rapid response times [39].

## 5. Case Studies

### 5.1 Multi-Core Processing in Industrial Robots

Industrial robots are specifically engineered to perform repetitive and precise tasks, frequently in settings that are hazardous for humans. They assemble components, wield materials, coat surfaces, and handle materials. Traditional single core processors struggle to handle the concurrent processing needs of industrial robots to perform these tasks. Multi-core processing has become an essential technology in addressing these requirements, allowing robots to carry out multiple activities concurrently and with greater efficiency.

#### 5.1.1 Case Study 1

ABB Robotics, a leading manufacturer of industrial robots, has integrated multi-core processors into its latest robotic systems [40].

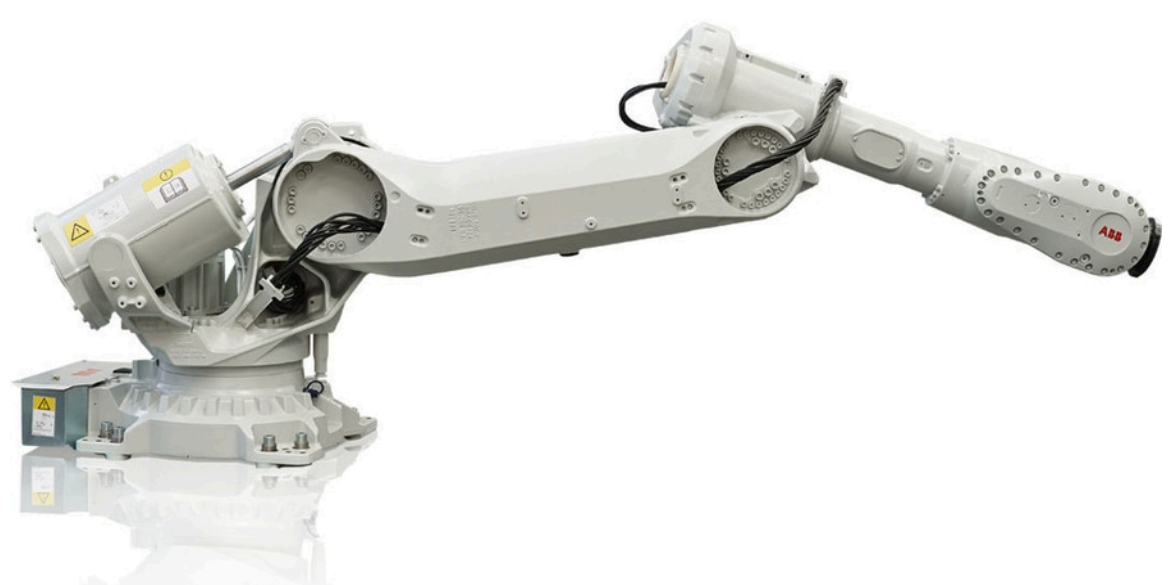

*Figure 5: ABB Robotics, IRB 6700 Series*

**Company/Project Name:** ABB Robotics, IRB 6700 Series.
**Description of the Robot:** The IRB 6700 series is designed for heavy-duty tasks such as spot welding, material handling, and machine tending.
**Technical Specifications:** Equipped with a quad-core processor, each running at 2.5 GHz, the IRB 6700 series can handle complex computations and multiple data streams simultaneously.
**Performance Outcomes:** The multi-core processor enables the IRB 6700 to perform tasks 20% faster than previous models with single-core processors. Additionally, there has been a 15% decrease in energy consumption, and the robot's accuracy has increased by 10%.

### 5.2 GPU Acceleration in Autonomous Vehicles

Autonomous vehicles rely on a suite of sensors, including cameras, LiDAR, radar, and ultrasonic sensors, to perceive their environment. They must process this data in real time to detect and recognize objects and road conditions, take decisions, and execute driving actions. GPU acceleration has become a crucial technology in meeting the computational demands of autonomous vehicles, leading to the adoption of GPUs for their superior parallel processing capabilities, as traditional CPU-based systems often struggle to meet these real-time demands.

#### 5.2.1 Case Study 2

NVIDIA, a leading technology company, has been at the forefront of integrating GPU acceleration into autonomous vehicles through their NVIDIA DRIVE platform [41].

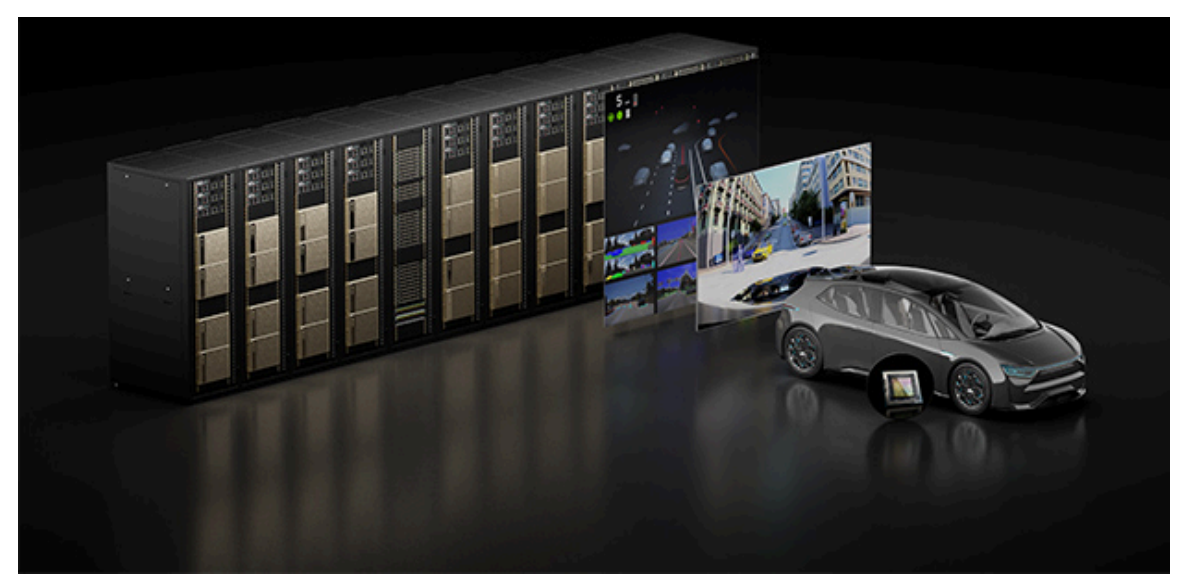

*Figure 6: NVIDIA DRIVE*

**Company/Project Name:** NVIDIA DRIVE.
**Description of the Vehicle:** Autonomous test vehicles equipped with the NVIDIA DRIVE platform, utilizing a range of sensors for perception and navigation.
**Technical Specifications:** The platform uses NVIDIA's powerful GPUs, such as the Xavier and Orin systems-on-a-chip (SoCs), featuring up to 12 cores and hundreds of tensor cores designed for AI computations.
**Performance Outcomes:** NVIDIA's test vehicles demonstrated significant improvements in object detection accuracy and processing speeds. For example, the vehicles achieved a frame rate of over 60 FPS for camera data processing, with a latency of less than 30 milliseconds, crucial for real-time decision-making.

### 5.3 FPGA Implementation in Real-Time Control Systems

Real-time control systems are designed to process inputs and generate outputs within a strict time frame, ensuring that the system responds predictably to external events. Industrial automation, advanced driver-assistance systems (ADAS), and unmanned aerial vehicles (UAVs) all use real-time control systems in addition to robotic arms and drones. Field-Programmable Gate Arrays (FPGAs) have become a key technology in meeting the stringent computational demands of these systems when traditional microprocessors and digital signal processors (DSPs) often struggle to meet these real-time requirements.

### 5.3.1 Case Study 3

Xilinx, a leading provider of FPGA technology, has implemented FPGA-based solutions in various real-time control systems [42].

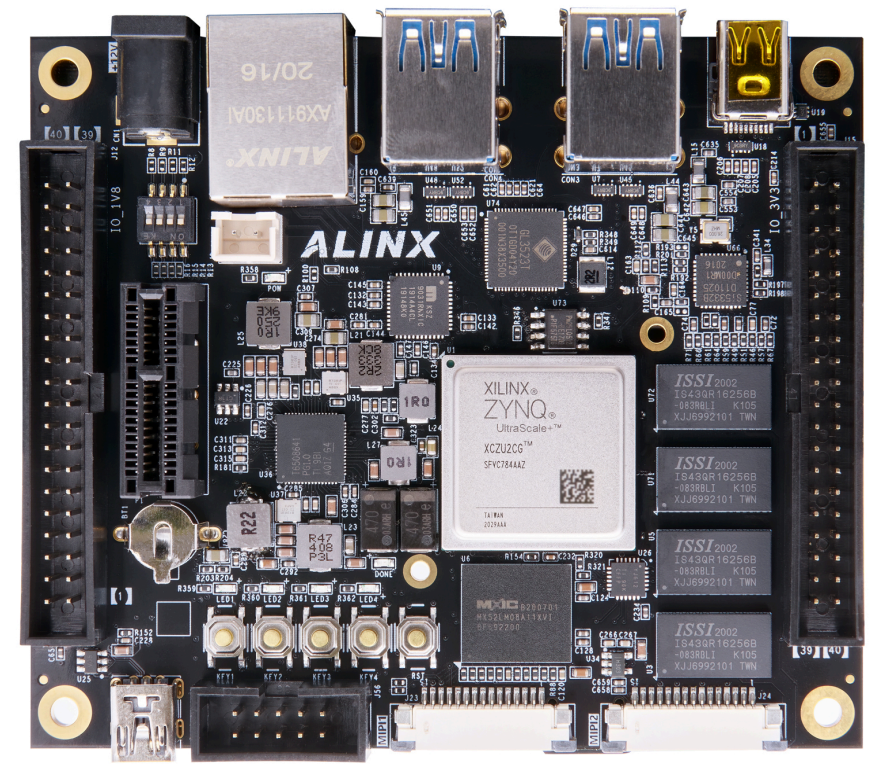

*Figure 7: Xilinx - Zynq UltraScale+ MPSoC*

**Company/Project Name:** Xilinx - Zynq UltraScale+ MPSoC in industrial motor control.
**Description of the System:** The Zynq UltraScale+ MPSoC integrates FPGA fabric with ARM processors, used in industrial motor control systems to manage precision motor operations in real time.
**Technical Specifications:** The system utilizes a Xilinx Zynq UltraScale+ MPSoC with up to 6 ARM Cortex-A53 cores, FPGA fabric with over 1 million logic cells, and high-speed I/O capabilities.
**Performance Outcomes:** Implementation of the FPGA-based motor control system resulted in a 50% reduction in control loop latency, improved motor precision by 30%, and reduced overall system power consumption by 20%.

### 5.4 Distributed Systems in Swarm Robotics

Swarm robots have applications in a variety of fields, including search and rescue, environmental monitoring, agricultural, and military operations. Swarm robots coordinate movement for particular area coverage and make decisions on the basis of local information. Swarm robotics involves the use of multiple robots working together to achieve a common goal. Swarm robotics aims to complete tasks through decentralized coordination and group behavior, drawing inspiration from the behavior of social insects like ants and bees, and distributed systems provide a framework for decentralized control and data sharing.

### 5.4.1 Case Study 4

Harvard University's Wyss Institute has developed a swarm robotics system called Kilobot, which demonstrates the power of distributed systems in coordinating large numbers of robots [43].

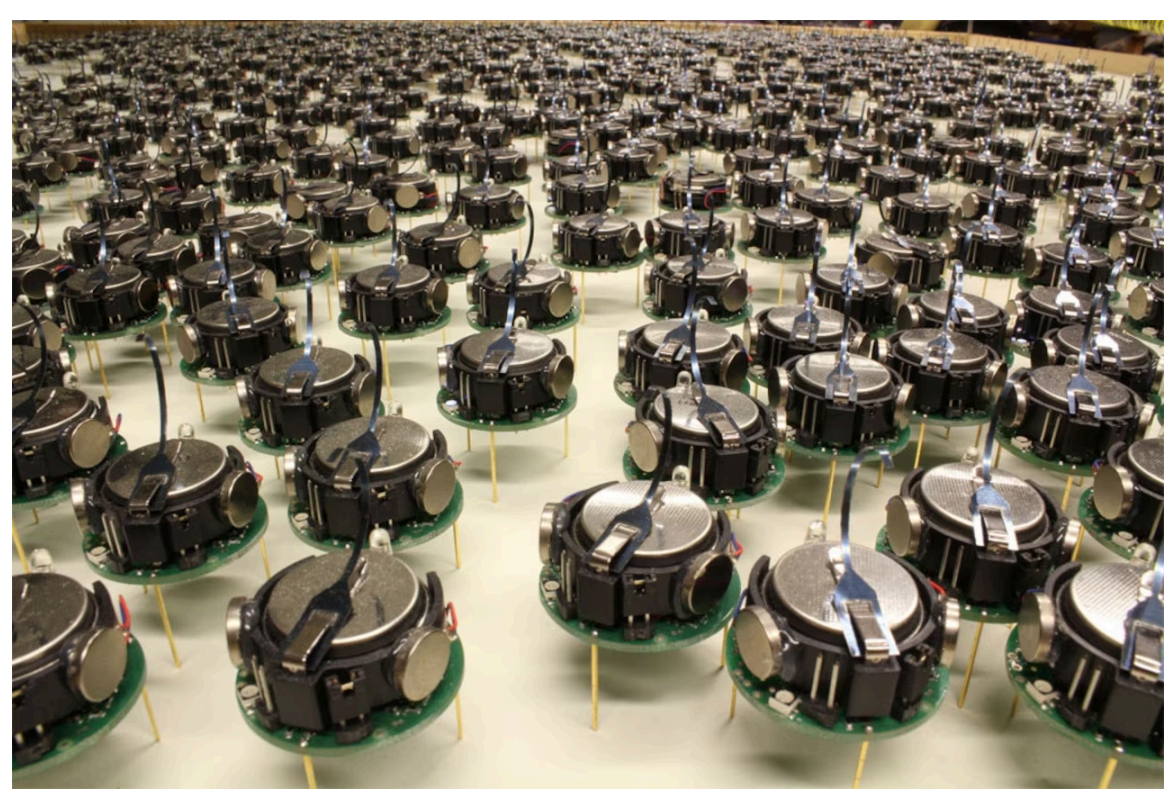

*Figure 8: Swarm Robots*

**Company/Project Name:** Harvard Wyss Institute - Kilobot Project.
**Description of the Swarm:** Kilobots are small, low-cost robots designed to work in large swarms. They perform tasks such as shape formation, collective transport, and environmental monitoring.
**Technical Specifications:** Each Kilobot uses simple infrared communication for local interactions, a vibrating motor for movement, and a basic microcontroller for processing. The swarm operates using decentralized algorithms that govern behavior based on local information.
**Performance Outcomes:** The Kilobot project demonstrated the ability of distributed systems to achieve complex behaviors with minimal communication overhead. For instance, the robots can form specific shapes and patterns with high accuracy and robustness. Task completion rates improved significantly compared to traditional centralized approaches.

**6. Architecture Comparison**

**6.1 Speed and Efficiency Comparison**

| Architecture | Speed | Efficiency | Case Example |
|---|---|---|---|
| Multi-Core Processors | **Parallel Task Execution**: Execute multiple tasks concurrently, improving processing speed and reducing latency. **Load Balancing**: Effective load balancing prevents bottlenecks, maintaining high speed. | **Resource Utilization:** High efficiency through balanced task distribution across cores. | **Industrial Robots**: Multi-core processors in manufacturing robots show improved task execution speed and efficiency. |
| Graphics Processing Units (GPUs) | **High Throughput Computing**: Handle thousands of parallel threads, ideal for data-intensive tasks. **Real-Time Image Processing**: Enhance speed and accuracy in robotic vision applications. | **Data Processing**: Efficient for high-throughput computing and data-heavy applications. | **Autonomous Vehicles**: NVIDIA GPUs enable real-time processing of camera and LiDAR data for autonomous driving. |
| Field-Programmable Gate Arrays (FPGAs) | **Customized Parallelism**: Execute specific tasks in parallel, tailored to application requirements. **Low Latency**: Provide low-latency processing for immediate response times. | **Task-Specific Efficiency**: Highly efficient for specialized tasks due to customizability. | **Robotic Control Systems**: FPGAs in control systems process sensor inputs and commands with high speed and low latency. |
| Distributed Systems and Cloud Computing | **Scalability**: Scale to accommodate more robots and complex tasks, enhancing speed and capacity. **Data Distribution**: Efficiently manage data distribution and aggregation. | **Resource Sharing**: High efficiency through parallel processing and load distribution across multiple nodes. | **Swarm Robotics**: Distributed systems improve task completion rates and robustness through efficient data sharing. |

**6.2 Scalability Comparison**

| Architecture | Scalability Factors | Case Example |
|---|---|---|
| Multi-Core Processors | **Core Addition**: Adding more cores generally improves performance linearly until memory bandwidth becomes a bottleneck. **Software Parallelism**: Software must be designed to take advantage of multiple cores, which can be challenging for some algorithms. | **Industrial Automation**: Multi-core processors in industrial robots, such as assembly line robots, show improved performance as more cores are added, provided the software is optimized for parallelism. |
| Graphics Processing Units (GPUs) | **Thread Management**: GPUs can manage thousands of threads concurrently, making them highly scalable for parallel tasks. **Data Parallelism**: Tasks that can be broken down into smaller, independent operations scale well with | **Autonomous Vehicles:** NVIDIA GPUs in autonomous vehicles show excellent scalability in processing larger datasets from cameras and LiDAR as more GPU resources are |

|  | additional GPU resources. | added. |
|---|---|---|
| Field-Programmable Gate Arrays (FPGAs) | **Configurable Logic Blocks**: FPGAs scale by adding more configurable logic blocks, but scalability is task-specific.<br>**Task-Specific Scaling**: The scalability of FPGAs depends on how well the task can be parallelized and mapped onto the FPGA architecture. | **Robotic Control Systems**: FPGAs in robotic control systems demonstrate good scalability for specific tasks like sensor processing, where additional logic blocks can be configured for parallel processing. |
| Distributed Systems and Cloud Computing | **Node Addition**: Adding more nodes to a distributed system generally improves scalability linearly if the communication overhead is managed effectively.<br>**Distributed Algorithms**: Algorithms must be designed to minimize inter-node communication and maximize parallel processing. | **Swarm Robotics**: Distributed systems in swarm robotics, such as drone fleets, show high scalability as more nodes are added, enabling more complex and coordinated behaviors. |

### 6.3 Energy Consumption Comparison

| **Architecture** | **Energy Consumption Factors** | **Case Example** |
|---|---|---|
| Multi-Core Processors | **Dynamic Voltage and Frequency Scaling (DVFS)**: Adjusting voltage and frequency based on workload reduces energy consumption.<br>**Core Utilization**: Efficient utilization of cores minimizes idle power draw. | **Robotic Arms**: Multi-core processors in robotic arms use DVFS to manage energy consumption during varied workloads. |
| Graphics Processing Units (GPUs) | **High-Performance Computing**: GPUs consume significant power during intensive computations, but offer high power efficiency for parallel tasks.<br>**Power Management Techniques**: Techniques like power capping and load balancing help optimize energy usage. | **Autonomous Drones**: NVIDIA GPUs in drones use power management techniques to balance performance and energy consumption. |
| Field-Programmable Gate Arrays (FPGAs) | **Customized Configurations**: Task-specific configurations can be highly energy efficient.<br>**Low-Power Modes**: Utilizing low-power modes reduces energy consumption during idle periods. | **Real-Time Control Systems**: FPGAs in real-time control systems use customized configurations to achieve energy-efficient processing. |
| Distributed Systems and Cloud Computing | **Resource Allocation**: Efficient allocation of resources across distributed systems minimizes energy usage.<br>**Data Center Energy Management**: Techniques such as server consolidation and cooling optimization reduce energy consumption in data centers. | **Swarm Robotics**: Distributed systems in swarm robotics use efficient resource allocation to manage energy consumption across multiple robots. |

### 6.4 Summary Table of Architecture Comparison

| **Architecture** | **Performance** | **Scalability** | **Energy Consumption** | **Flexibility** | **Cost** |
|---|---|---|---|---|---|
| Multi-Core Processors | Good parallel task execution | Moderate | Efficient with DVFS | Versatile for various applications | Generally cost-effective |

| Graphics Processing Units (GPUs) | High throughput for parallel processing | Excellent | High power consumption, efficient for large-scale tasks | Suitable for image processing and ML | Higher initial cost, cost-effective for specific tasks |
|---|---|---|---|---|---|
| Field-Programmable Gate Arrays (FPGAs) | High performance for specialized tasks | Highly scalable for specific applications | Very energy-efficient for custom configurations | Highly flexible but design-intensive | Higher initial cost, cost-effective long-term for specialized use |
| Distributed Systems and Cloud Computing | High performance for large-scale tasks | Virtually unlimited | Varies, benefits from resource optimization | Extremely flexible | Variable, potentially high operational costs |

## 7. Challenges and Potential Solutions

### 7.1 Hardware Limitations

Although there have been improvements in multi-core CPUs, GPUs, and FPGAs, there are still constraints on the sheer processing power required for complex robotic tasks. For battery-powered robotic applications, finding a balance between performance and power drain has been a consistent challenge. High-performing computational units generate massive heat, necessitating thermal management for optimal performance [44].

### 7.2 Software Challenges

Efficient parallel code writing is a challenging task, as developers need to focus on concurrency, synchronization, and data dependencies. Achieving real-time performance, ensuring data consistency, and synchronizing nodes where data discrepancies can occur can prove to be a very difficult task. Moreover, debugging parallel applications is more complex than debugging sequential programs, as errors may be intermittent and difficult to reproduce [45].

### 7.3 Integration Challenges

Different parallel computing architectures may have compatibility issues, and integrating them into a single robotic system can prove to be difficult. It is a significant challenge to ensure smooth communication and operation between heterogeneous components, reduce communication latency, and develop hardware and software in tandem to optimize performance [46].

### 7.4 Environmental and Operational Challenges

Operating in dynamic and unpredictable environments requires robust computing solutions to ensure reliable performance, which is challenging. Mobile robotics often faces constraints on computational resources, power, and memory. Furthermore, in critical applications such as healthcare and autonomous driving, ensuring the fail-safe mechanisms, error handling, safety, and reliability of robotic systems is paramount and a difficult task [47].

### 7.5 Potential Solutions and Innovations

New multi-core processors are being developed with better performance, and combining various types of processors, like CPUs, GPUs, and FPGAs, in a system is allowing us to take advantage of each type, boosting overall performance. Dynamic power management methods like adaptive voltage scaling and power gating can make systems energy efficient. Liquid and thermoelectric cooling can aid in managing overheating. The development of more accessible parallel programming models and enhancing real-time operating systems (RTOS) can address some of the software challenges. Lock-free algorithms and transactional memory can be used for data consistency and reduced latency. Standard interface and protocol development, innovation in communication protocols, and co-design methodologies that optimize the interaction between hardware and software can all help to address integration problems. For better decision-making, machine learning and AI are useful, allowing robotic systems to learn from their experiences [48, 49, 50]. The development of robust computer architectures, efficient management of computational resources, and innovative solutions to fail-safe mechanisms can tackle various operational challenges [51, 52, 53].

### 7.6 Summary Table of Challenges and Potential Solutions

| **Challenge** | **Multi-Core Processors** | **GPUs** | **FPGAs** | **Distributed Systems** | **Potential Solutions** |
|---|---|---|---|---|---|
| Processing Power | Limited scalability for very high-performance tasks | High performance but high power consumption | High performance for specific tasks | Dependent on network and resource allocation | Advanced processor designs, heterogeneous computing |
| Energy Efficiency | Efficient with DVFS but still power-hungry | High power consumption | Very efficient for specific configurations | Varies based on data center efficiency | Energy-efficient architectures, dynamic power management |
| Thermal Management | Challenges in high-performance scenarios | Significant heat generation | Efficient but requires good cooling design | Data center cooling efficiency | Advanced cooling technologies, thermoelectric cooling |
| Parallel Programming | Complex, requires specialized knowledge | Complex, especially for non-graphics tasks | Requires detailed hardware knowledge | Complex, especially for large-scale systems | High-level parallel programming models, improved synchronization mechanisms |
| Real-Time Processing | Achievable but challenging | Difficult due to high latency | Excellent for specific tasks | Challenging due to network latency | Enhancements in RTOS, low-latency communication protocols |
| Synchronization | Moderate difficulty | High difficulty due to many threads | Moderate difficulty | Very challenging | Advanced synchronization techniques, improved profiling tools |
| Integration | Compatibility issues with other architectures | Compatibility and latency issues | Compatibility and design complexity | Compatibility and latency issues | Standardized interfaces and protocols, hardware-software co-design methodologies |
| Communication Latency | Moderate latency | High latency | Low latency | High latency | Low-latency communication protocols |
| Dynamic Environments | Moderate adaptability | Moderate adaptability | High adaptability for specific tasks | High adaptability | Robust and resilient designs, adaptive algorithms |

| Safety and Reliability | Moderate reliability | Moderate reliability | High reliability for specific tasks | High reliability | Fail-safe mechanisms, redundant systems |
|---|---|---|---|---|---|

**8. Future Research Directions**

| Research Area | Future Research Directions |
|---|---|
| Processor Architectures | Next-generation multi-core processors, quantum computing, and neuromorphic computing |
| Parallel Programming Models | High-level abstractions, domain-specific languages, automated parallelization |
| Real-Time Processing | Real-time algorithms, latency reduction techniques, and predictive processing |
| Integration Techniques | Seamless hardware-software integration, modular architectures, and interoperability standards |
| Adaptive Systems | Adaptive computing architectures, intelligent resource management, and learning-based optimization |
| Sustainability | Adaptive computing architectures, intelligent resource management, and learning-based optimization |
| Security and Reliability | Robust security protocols, fault-tolerant architectures, redundancy and backup systems |

**9. Discussion**

Robotics makes use of multicore processors due to their versatility and excellent performance balance. GPUs have advanced a lot throughout the years, and they are ideal for parallel processing, but there is the issue of power consumption and heat dissipation. FPGAs offer high-performance solutions with low latency, but the downside is the requirement of specialized programming knowledge and a long development period. Distributed systems enable large-scale programming while introducing challenges related to synchronization and security. Case studies demonstrate how multi-core processing, GPU acceleration, FPGAs, and distributed systems enhance robotic applications in industrial robotics, autonomous vehicles, real-time control systems, and swarm robotics. Various parallel computing architectures have advanced robotics applications, introducing hardware, software, integration and operational challenges. By conducting thorough research and implementing innovative ideas, we can overcome these obstacles, leading to further advancements in robotics fields.

**10. Conclusion**

Parallel computing architectures in robotics have the potential to revolutionize and advance robotic fields and applications in terms of performance enhancement, energy efficiency, scalability, and advanced functionality. Despite the significant progress, the complexity of robotic applications continues to increase, prompting the evolution of parallel computing architectures to meet these needs. Continued research in this field is required to explore new hybrid architectures, improve software development for user-friendly parallel programming models and hardware-software co-design and integration methods, and enhance the security of parallel computing architectures for defense against cyber security threats. By furthering the progress of parallel computing architectures, more potent, efficient, and intelligent systems can be created that hold the capacity to transform various industries.